\documentclass{IEEEtran}

\usepackage{microtype}
\usepackage{graphicx}
\usepackage{amsmath,amssymb,bm}
\usepackage{subfigure}
\usepackage{booktabs} 
\usepackage{hyperref}

\usepackage{makecell}
\usepackage{siunitx}
\usepackage[numbers]{natbib}
\DeclareMathOperator{\E}{\mathbb{E}}

\author{Matthew Amodio \\ matthew.amodio@yale.edu \and Smita Krishnaswamy \\ smita.krishnaswamy@yale.edu}

\title{MAGAN: Aligning Biological Manifolds}
\begin{document}

\maketitle




\begin{abstract}
It is increasingly common in many types of natural and physical systems (especially biological systems) to have different types of measurements performed on the same underlying system. In such settings, it is important to align the manifolds arising from each measurement in order to integrate such data and gain an improved picture of the system. We tackle this problem using generative adversarial networks (GANs). Recently, GANs have been utilized to try to find correspondences between sets of samples. However, these GANs are not explicitly designed for proper alignment of manifolds. We present a new GAN called the Manifold-Aligning GAN (MAGAN) that aligns two manifolds such that related points in each measurement space are aligned together. We demonstrate applications of MAGAN in single-cell biology in integrating two different measurement types together. In our demonstrated examples, cells from the same tissue are measured with both genomic (single-cell RNA-sequencing) and proteomic (mass cytometry) technologies. We show that the MAGAN successfully aligns them such that known correlations between measured markers are improved compared to other recently proposed models.
\end{abstract}

\section{Introduction}
We commonly face the situation of having samples from a pair of related domains and want to ask the natural question of how samples from one relate to samples from the other. Our motivational system for this is two types of measurements on cells sampled from the same population in a biological system. It is important for the discovery of new biology to integrate these datasets, which are often generated at great cost and expense. However, a fundamental challenge is that there are exponentially many possible relationships that could exist between the two domains of measurement and the system must learn a logical way to map between them.

The first approaches for teaching neural networks to learn these relationships required supervised paired examples from each domain, an impractical demand for many applications \cite{isola2016image}. Recently, there have been attempts at performing the same task without the supervision of paired data \cite{zhu2017unpaired,yi2017dualgan,kim2017learning}. Like these previous models, the MAGAN learns to map between distinct domains from unsupervised, unpaired data without pretraining. It can take a point in the first domain and generate a point that is indistinguishable from points in the other domain. However, unlike previous models, the MAGAN learns the most \textit{coherent} mapping, rather than an arbitrary one. The MAGAN will not just take a point in the first domain and generate any point from the second domain, but it will generate the most closely related one. This is achieved by aligning, rather than superimposing, the manifolds of the two domains.

\begin{figure}
\centering
\includegraphics[width=.5\textwidth]{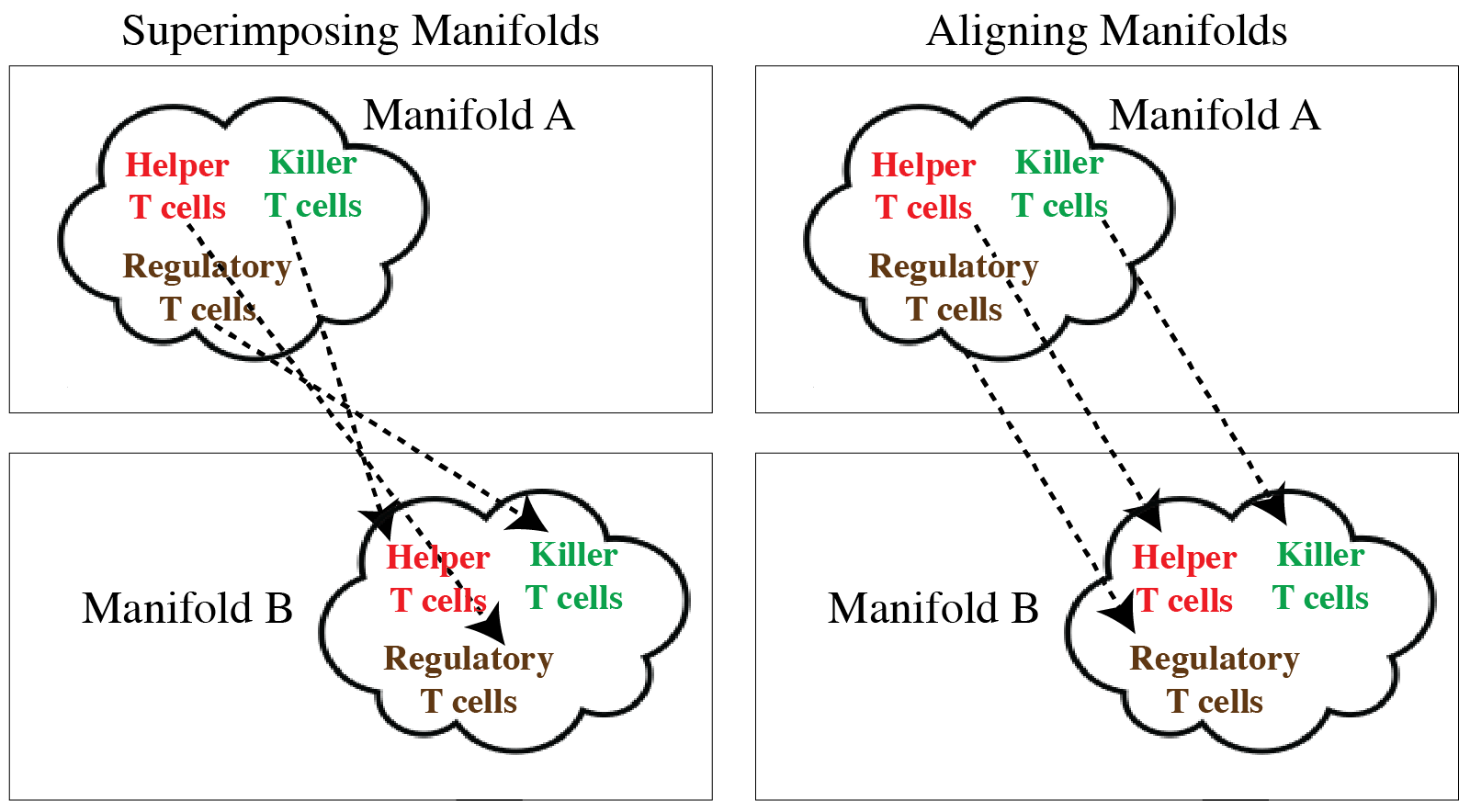}
\caption{There are exponentially many mappings that superimpose the two manifolds, fooling a GAN's discriminator. By aligning the manifolds, we maintain pointwise correspondences.}
\label{fig:cartoon}
\end{figure}

\begin{figure*}
\centering
\includegraphics[width=.6\textwidth]{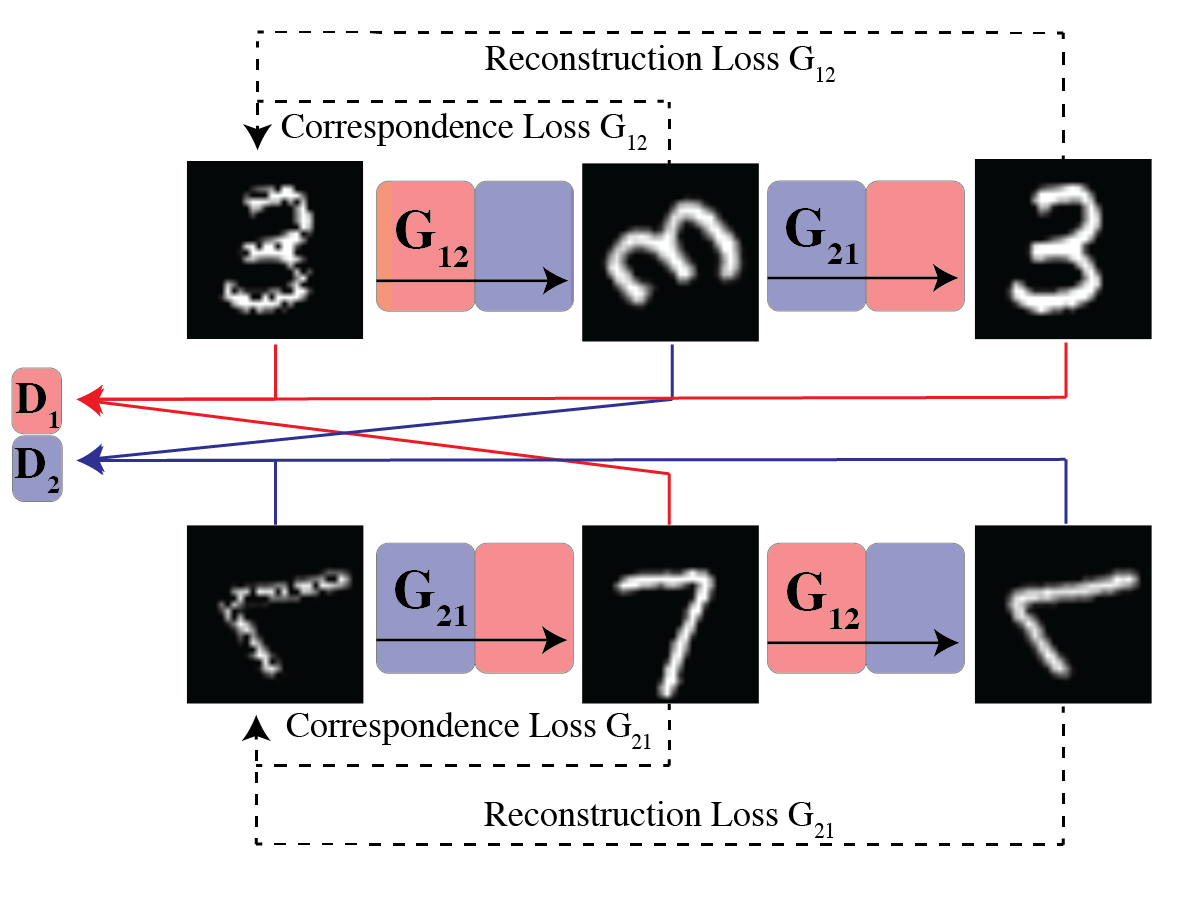}
\caption{The MAGAN architecture with two generators, two discriminators, reconstruction loss, and correspondence loss. Domain 1 comprises upright images of 3's and 7's, Domain 2 comprises rotated images of 3's and 7's.}
\label{fig:architecture}
\end{figure*}

The high-dimensional inputs that are typical for neural network applications can typically be modeled very well with a lower-dimensional manifold \cite{domingos2012few}. Much work has framed the generation problem of GANs as sampling points from this manifold \cite{park2017mmgan,zhu2016generative}. Here, each domain lies on a manifold and we want to find an alignment between them.

We first consider an example of the difference between superimposing and aligning manifolds on image domains. Earlier work \cite{kim2017learning} has demonstrated that an image of an object in the first domain can be mapped to an object of another object in the second domain while preserving the orientation with respect to the picture frame. However, in those cases, the orientation axis can be completely reversed. An image in the first domain facing \ang{30} maps to an image in the second domain facing \ang{150}, and vice versa. The mappings successfully fool the discriminator in each domain at the level of an entire batch (the manifolds are superimposed), but there are other mappings that also fool the discriminator that preserve the individual pointwise structure of the original domain. Namely, the optimal alignment would map first domain images at \ang{30} to second domain images also at \ang{30}. Without aligning the manifolds, only random initialization will determine which superimposition is learned on any particular attempt.

While the preference for the logical mappings of manifold alignment over the arbitrary ones of manifold superimposition is of general interest to all domains, we present multiple applications in single-cell biological analysis where it is essential. We propose the novel concept of using adversarial neural networks for alignment of manifolds arising from different biological experimental data measurement types.  

Single-cell biological experiments create many situations where manifold alignment problems are of interest. New technologies allow for measurements to be made at the granularity of each cell, rather than older technologies which could only acquire aggregate summary statistics for whole populations of cells. While these instruments allow us to discover biological phenomena that were not apparent before, it is a challenge to integrate and analyze this information in a unified fashion for biological discovery. Further, even for the same technology, experiments run on different days or in different batches can show variations even on the same populations, possibly due to calibration differences. In such cases even replicate experiments need alignment before comparison. Two such technologies that we examine are single-cell RNA sequencing which measures cells in thousands of gene (mRNA) dimensions and mass cytometry which measures protein abundances in several dozen dimensions \cite{bendall2012deep, klein2015droplet}. 

In all of these examples we have two data manifolds with a latent physical cell being measured analogously in each manifold. In some applications it might be adequate to simply superimpose these manifolds in any way. In many applications though, including the ones demonstrated here, we would like to be able to align them such that the two representations of each latent cell are aligned. The MAGAN presented here improves upon neural models for manifold alignment by finding the mapping between the manifolds (\textit{correspondence}) that models these latent points by penalizing differences in each point's representation in the two manifolds.

We summarize the contributions of this paper as follows:
\begin{enumerate}
\item The introduction of a novel GAN architecture that aligns rather than superimposes manifolds to find relationships between points in two distinct domains
\item The demonstration of novel applications made possible by the new architecture in the analysis of single-cell biological data
\end{enumerate}

The rest of this paper is organized as follows. First, there is a detailed description of the MAGAN architecture. Next, there is a validation of its performance on artificial data and the standard MNIST dataset. Then, there are demonstrations on three real-world biological applications: mapping between two replicate cytometry domains, mapping between two different cytometry domains, and mapping between one cytometry domain and a single-cell RNA sequencing domain.

\section{Model}
In this section we detail the MAGAN architecture and specify the notation used thereafter. We then elaborate on the key novel aspects of the model individually, discussing in turn the unsupervised correspondence loss, semi-supervised correspondence loss, and data augmentation.

\subsection{Architecture}
The MAGAN (Figure \ref{fig:architecture}) is composed of two GANs, each with a generator network $G$ that takes as input $X$ and outputs a target dataset $X^{\prime}$. We refer to each generator as a \textit{mapping} from the input domain to the output domain. Each generator attempts to make its output $G(X)$ indistinguishable by $D$ from $X^{\prime}$. Denote the two datasets $X_1$ and $X_2$. Let the generator mapping from $X_1$ to $X_2$ be $G_{12}$ and the generator mapping from $X_2$ to $X_1$ be $G_{21}$. The discriminator that tries to separate true points from mapped ones for the first domain is $D_1$ and the discriminator doing so for the second domain is $D_2$.

The loss for $G_1$ on minibatches $x_1$ and $x_2$ is:
\begin{gather}
x_{12} = \bm{G_{12}}(x_1) \nonumber \\
x_{121} = \bm{G_{21}}(x_{12}) \nonumber \\
L_r = L_{reconstruction} = L(x_1, x_{121}) \nonumber \\
L_d = L_{discriminator} = - \E_{x_1  \sim P_{X_1}} \left [ log \bm{D_2}(x_{12}) \right] \nonumber \\
L_c = L_{correspondence} = L(x_1, x_{12}) \nonumber \\
\bm{L_{\bm{G_1}}} = L_{r} + L_{d} + L_{c} \nonumber
\end{gather}
where $L$ is any loss function, here mean-squared error (MSE). 

Similarly, the loss for $G_2$ is:
\begin{gather}
x_{21} = \bm{G_{21}}(x_2) \nonumber \\
x_{212} = \bm{G_{12}}(x_{21}) \nonumber \\
L_r = L(x_2, x_{212}) \nonumber \\
L_d = - \E_{x_2  \sim P_{X_2}} \left [ log \bm{D_1}(x_{21}) \right ] \nonumber \\
L_c = L(x_2, x_{21}) \nonumber \\
\bm{L_{\bm{G_2}}} = L_{r} + L_{d} + L_{c} \nonumber
\end{gather}

The losses for $D_1$ and $D_2$ are:
\begin{gather}
\bm{L_{\bm{D_1}}} = - \E_{x_1  \sim P_{X_1}} \left [ log \bm{D_1}(x_1) - log \bm{D_1}(x_{121}) \right ] \nonumber \\
- \E_{x_2  \sim P_{X_2}} \left [ log (1 - \bm{D_1}(x_{21})) \right ] \nonumber \\
\bm{L_{\bm{D_2}}} = - \E_{x_2  \sim P_{X_2}} \left [ log \bm{D_2}(x_2) - log \bm{D_2}(x_{212}) \right ] \nonumber \\
- \E_{x_1  \sim P_{X_1}} \left [ log (1 - \bm{D_2}(x_{12})) \right ] \nonumber
\end{gather}

A crucial implementation decision is to tie the weights of $G_{12}$ and $G_{21}$ for both directions of the mapping, as this ensures that the mappings will be between similar data points. For example, after mapping a point from $X_1$ to $X_2$, in order to reconstruct the original point $x_1$, the first mapping must preserve enough information in $x_{12}$. Then, since the weights are tied, $G_{21}$ must use this information in the same way when asked to map an original point $x_2$.

\subsection{Correspondence Loss}
\subsubsection{Unsupervised Correspondence}
Previous models included only two restrictions: (1) that the two generators be able to reconstruct a point after it moves to the other domain and back, and (2) that the discriminators not be able to distinguish batches of true and mapped points. To do this, the generators could learn arbitrarily complex mappings as long as they superimpose the two manifolds.

To instead enforce the manifolds be fully aligned, the MAGAN includes a correspondence loss between a point in its original domain and that point's representation after being mapped to the other domain. This correspondence loss can be chosen appropriately for the manifolds in any particular problem. In the biological domains considered here, there are a subset of shared features measured in both experiments. We use the MSE over these subsets as the correspondence loss, as we do not want to reorder sets of points or change their representations more than what is required to match the other domain. 

\begin{figure*}
\centering
\includegraphics[width=.8\textwidth]{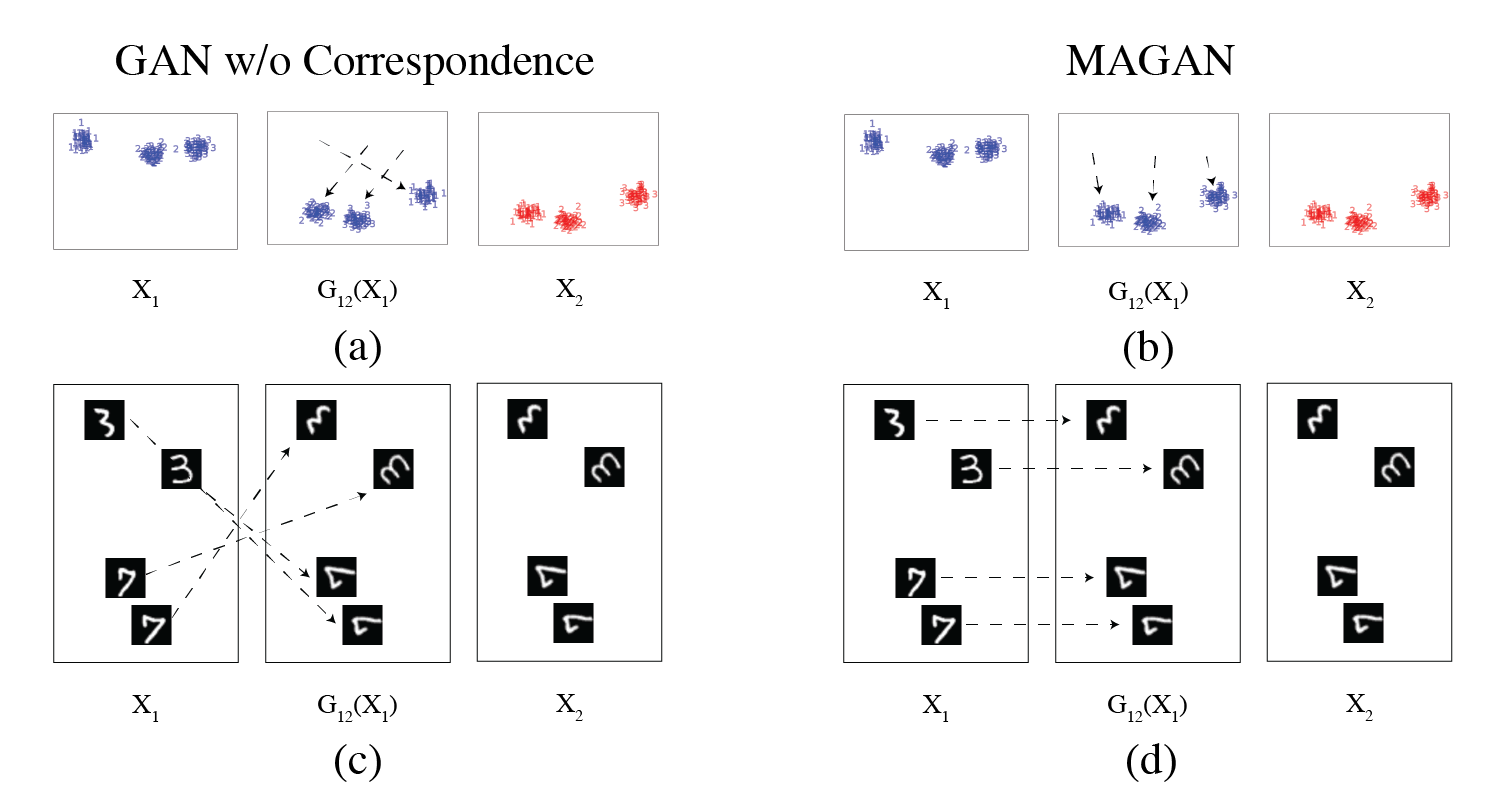}
\caption{Both models superimpose the manifolds, meaning the first domain ($X_1$) is mapped to the second domain ($X_2$) such that the dataset of the first domain after mapping ($G_{12}(X_1$)) matches the second domain. Without the correspondence loss, though, this mapping is arbitrary and thus the relationships found vary. With the correspondence loss, the relationships found are coherent. This is confirmed with (a) a GAN without correspondence loss on artificial data (b) MAGAN on artificial data (c) a GAN without correspondence loss on MNIST and (d) MAGAN on MNIST.}
\label{fig:example_mappings}
\end{figure*}

\subsubsection{Semi-supervised Correspondence}
The MAGAN's correspondence loss additionally provides a natural opportunity for semi-supervised learning. If each point in $X_1$ already had a known correspondence with a point in $X_2$, no framework of dual GANs would be necessary to discover relationships. In some domains, though, it is difficult to define a meaningful distance measure but easy to acquire a very small number of labeled pairs. We would like a model that learns from unsupervised data but can improve with any small number of labels that can be acquired. In those situations, we want to leverage both (1) the information that the unsupervised model has learned on all of the data and (2) incorporate the information the labels provide where they exist.

The MAGAN can be used in this setting without any further modification using the following choice of correspondence loss function. We choose the loss function to be nonzero only at the paired points in each domain. Its value is then the sum of the losses on each labeled pair, where the loss for a particular labeled pair $(x_{1i},x_{2j}), x_{1i} \in X_1, x_{2j} \in X_2$ is:
\begin{gather}
L_{c} = MSE(\bm{G_{12}}(x_{1i}), x_{2j}) + MSE(\bm{G_{21}}(x_{2j}), x_{1i}) \nonumber
\end{gather}

\subsection{Manifold Data Augmentation}
The MAGAN utilizes a novel technique for data augmentation, leveraging the imperfect reconstructions each generator produces within its domain. It has been well established that autoencoders model and reconstruct from the data manifold \cite{hinton1997modeling,vincent2008extracting}. We note that the dual GANs within each domain function as an autoencoder, meaning their reconstruction $x_i^{\prime}$ of a sample $x_i$ is another point near the underlying manifold, but importantly $x_i^{\prime} \ne x_i$. By letting each discriminator see the reconstructions as true samples from the real domain, we both (1) augment the original data with new samples from the manifold and (2) prevent the discriminators from learning to separate real from generated examples by modeling the noise around the manifold, which differs between $X_1$ and $G_{21}(X_2)$ and between $X_2$ and $G_{12}(X_1)$. This is especially important in biological settings, where the number of measurements per cell dwarfs the number of cells measured and dropout in the measuring process produces sparsity.

\section{Experiments}

All experiments were performed with the MAGAN framework with discriminators of five layers each and generators of three layers each. Layer sizes depended on the dataset, while Leaky ReLU activations were used on all layers except the output layers of the discriminators (which were sigmoid) and the generators (which were linear). Dropout of 0.9 was applied during training and for images convolutional layers were used. Optimization was performed on 100,000 iterations of batches of size 256 by the ADAM optimizer with learning rate 0.001.

As with other GANs, the generators and discriminators are trained alternatively, so they each must get progressively better as their adversaries make their tasks harder and harder. One known difficulty in the adversarial training process is preventing a collapse of the generator into mapping all inputs to one point, chasing the minimum probability region of the discriminator as it moves. To combat this, the MAGAN includes the approach outlined in \cite{salimans2016improved}. This involves giving the discriminator access to minibatch information by having a subset of the network process a rotation of the original data matrix.

\subsection{Artificial Data}
We first test the MAGAN on a generated example of points sampled from Gaussian distributions with varying means. Figure \ref{fig:example_mappings}a shows the three subpopulations in the first domain $X_1$ in blue and the three in the second domain $X_2$ in red with an example mapping where, without the correspondence loss, each subpopulation in $X_1$ is mapped to a subpopulation in $X_2$, but not to the closest one. Even though the distribution of $G_{12}(X_1)$ matches the distribution of $X_2$, for an individual point $x_{1i} \in X_{1}$, $G_{12}(x_{1i})$ is not the member of $X_2$ that is most closely analogous to it. The MAGAN finds a mapping that fools the discriminator, too: the one that least alters the original input (Figure \ref{fig:example_mappings}b).

Without the correspondence loss, not only is a less-preferred manifold superimposition chosen, but the one chosen varies from run to run of the model. We compare the variability of the learned mappings across multiple runs of each model with 100 independent trials. In each trial we evaluate the relationships by calculating $G_{12}(x_{1i})$ for each $x_{1i} \in X_1$ and calculating its nearest neighbor $x_{2j}$ in the real $X_2$. Then, this is repeated for the other domain. Figure \ref{fig:simulations}a confirms that for the GAN without the correspondence loss, the learned manifold superimposition (and thus the correspondences) varies with repeated training the model. Figure \ref{fig:simulations}b confirms the MAGAN instead aligns the manifolds and finds the same correspondence every time.

\subsection{MNIST}
Next we test a subset of the MNIST handwritten digit data by taking only 3's and 7's as the first domain $X_1$, and a 120 degree rotation of each image as the second domain $X_2$. Without the correspondence loss (Figure \ref{fig:example_mappings}c), each subpopulation in $X_1$ maps to one of the subpopulations in $X_2$, but the original 3's go to the rotated 7's and vice versa. There is no term in the objective function to create a preference for the mapping that sends original 3's to rotated 3's. It would be difficult to define a distance measure that captures the notion of alignment with these manifolds, but it is a natural place where a small number of labeled pairs could be easily acquired. The semi-supervised correspondence loss with just a single labeled pair of points finds the desired manifold alignment and gets the correct correspondences for all of the other points that are unlabeled (Figure \ref{fig:example_mappings}d).

\begin{figure}
\centering
\includegraphics[width=.5\textwidth]{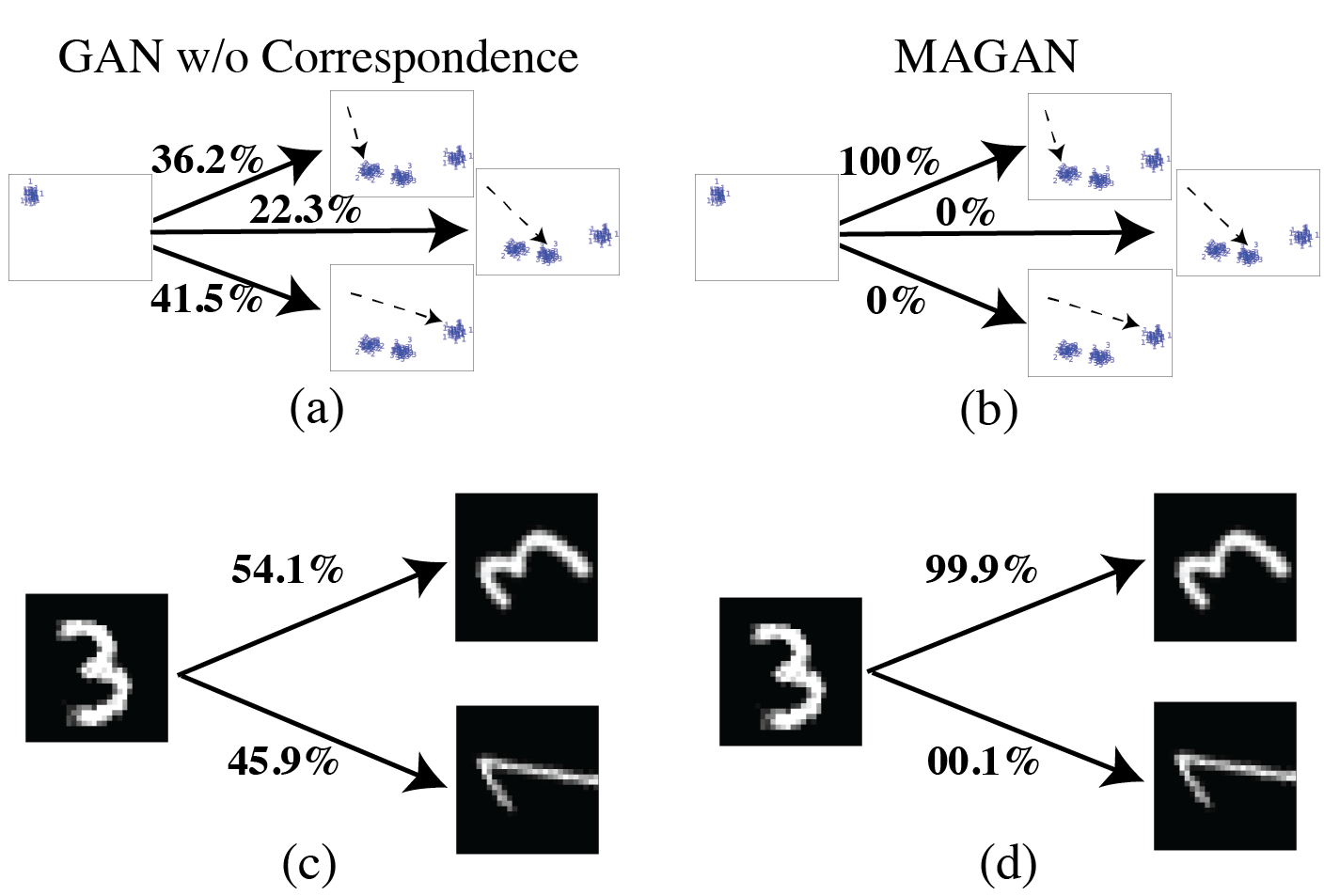}
\caption{In simulations of 100 complete training runs of each model, without correspondence loss the resulting relationships learned varied randomly in both the (a) toy and (c) MNIST datasets. With correspondence loss, the most coherent relationship was found repeatedly for both (b) toy and (d) MNIST datasets.}
\label{fig:simulations}
\end{figure}

Using the same simulation design as in the previous section, we can test the robustness of the models in finding these particular mappings. The GAN without the correspondence loss discovers either relationship with roughly even probability (Figure \ref{fig:simulations}c). Remarkably, the MAGAN is able to use the single labeled example to learn that (except for a few sloppily written 3's that in fact look more like 7's) the original 3's correspond to the rotated 3's and that the original 7's correspond to the rotated 7's every time (Figure \ref{fig:simulations}d).

\begin{figure}
\centering
\includegraphics[width=.5\textwidth]{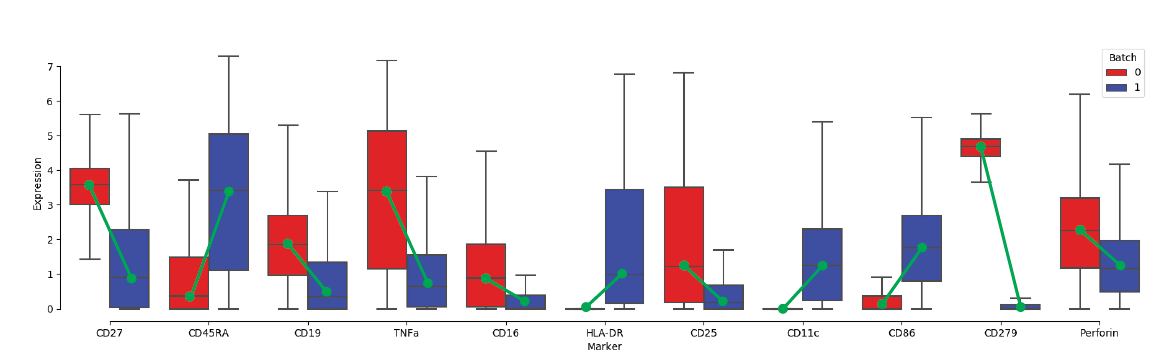}
\caption{Selected markers illustrating large batch effects that separate the two data manifolds.}
\label{fig:cytof_boxplots}
\end{figure}

\subsection{Correspondence Between CyTOF Replicates}
We now test the MAGAN on real biological data from single-cell time-of-flight mass cytometry (CyTOF) measurements of protein abundance. Each protein, also referred to as a \textit{marker}, is measured individually for each cell, allowing for more granular analysis than processes that only measure population totals for the cells in a given sample. Here the same sample was run twice in different batches (\textit{replicates}), but due to machine calibration and other experimental details that are impossible to reproduce precisely each time, there are distortions between the batches. Thus, even though the same physical blood sample is being measured, the data manifold of each batch is different. The type of noise introduced by these distortions is not known \textit{a priori}, need not fit any parametric assumption, and is likely to be highly nonlinear.

To analyze these two batches together, we need to know which cells in the first batch correspond to which cells in the second batch. To do this, we learn a mapping with the MAGAN between the batches, each of which contains 75,000 cells with 34 individual markers measured. Figure \ref{fig:cytof_boxplots} shows that the two batches indeed contain distinct differences in both the values of each marker and their distribution. For example, the mean value of HLA-DR in the second batch is higher than the \textit{maximum} value in the first batch.

\begin{figure}
\begin{minipage}[b]{.9\linewidth}
\centering
\includegraphics[width=\linewidth]{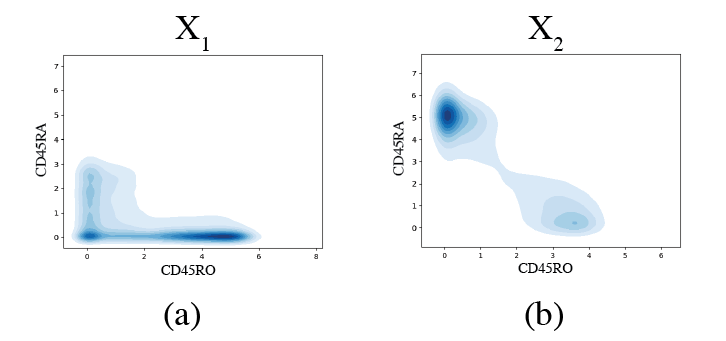}
\caption{Two distinct populations of T-cells (CD45RA+CD45RO- and CD45RA-CD45RO+) with severe dropout in the CD45RA marker that causes a difference between that between the (a) first batch and (b) second batch.}
\label{fig:cytof_tcells}
\end{minipage}

\begin{minipage}[b]{.9\linewidth}
\centering
\includegraphics[width=\linewidth]{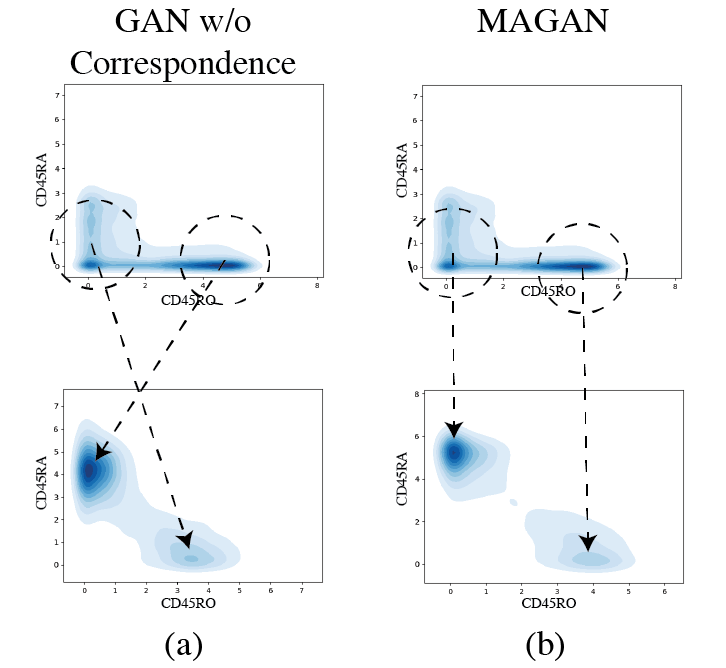}
\caption{(a) Without correspondence loss, the GAN corrects the batch effect but subpopulations are reversed. (b) The MAGAN still corrects the batch effect and subpopulations are preserved.}
\label{fig:cytof_tcells2}
\end{minipage}
\end{figure}

We demonstrate that the MAGAN with its correspondence loss preserves crucial information that is lost with the mapping from the GAN without the correspondence loss. Often, analysis starts by identifying subpopulations of interest. For example, naive T-cells and central memory T-cells serve distinct functions and can be identified by looking at two isoforms of the CD45 marker, CD45RA and CD45RO \cite{capra1999inmunobiology}. In naive T-cells, CD45RA is present while CD45RO is not (CD45RA+CD45RO-), and in central memory T-cells CD45RA is not present while CD45RO is (CD45RA-CD45RO+). Figure \ref{fig:cytof_tcells}a shows that very few cells had any CD45RA readings in the first batch, a typical case of instrument-induced dropout. Figure \ref{fig:cytof_tcells}b shows proper readings for CD45RA in the second batch, where the two distinct subpopulations are clearly seen.

Both models learn a mapping for the first batch of cells $x_1$ such that $G_{12}(x_1)$ fools their discriminators by looking like the second batch of cells $x_2$. However, in the GAN without the correspondence loss (Figure \ref{fig:cytof_tcells2}a), naive T-cells in the first batch are mapped to central memory T-cells in the second batch and vice versa. If we went through the manual process of gating (selecting cells by manually looking at relative marker expression) central memory T-cells in the first batch and wanted to know whether their expression was similar in the second batch, we would be led to believe incorrectly that either there are none of these cells in the second batch or their expression profile is radically different. 

The MAGAN learns a different mapping (Figure \ref{fig:cytof_tcells}b), the one in which subpopulation correspondences are preserved. Notably, the resulting mapped dataset $G_{12}(x_1)$ is not negatively affected by the correspondence loss. Instead, out of the two mappings that have similar results at the aggregate level, the one that maintains pointwise correspondences is learned. With the cell correspondences from other manifold superimpositions, the wrong biological conclusions could be made. This application necessitates the MAGAN's manifold alignment.

\subsection{Correspondence Between Different CyTOF Panels}
Next we demonstrate the MAGAN's ability to align two manifolds in domains whose dimensionality only partly overlap. Despite the other advantages of CyTOF instruments, one disadvantage is that CyTOF experiments can only measure the expression of 30-40 markers per cell. Each experiment chooses which 30-40 markers to measure and refers to this set as the \textit{panel}. Even though each panel has a limited capacity, different panels can be run on different samples from the same physical blood or tissue. The MAGAN provides the opportunity to combine the results from these multiple panels and effectively increase the number of expression measurements acquired for each cell.

To test this, we use the datasets from two experiments published in \cite{setty2016wishbone} where each experiment had a different panel that was run on samples from the same population of cells. The first panel measured 35 markers, the second panel measured 31 markers, and 16 of those were measured in both. Without any advanced methods, all we would be able to do across experiments is compare population summary statistics --- and lose all of the information at a single-cell resolution that motivated these experiments being done in the first place.

If we can identify points in each panel that measure the same cell, we can combine the measurements and have an augmented 50-dimensional dataset. To accomplish this, we take the first experiment's panel as one domain and the second experiment's panel as the other domain and use the MAGAN to learn a mapping between the two. We then combine the original 35 dimensions of a cell in the first experiment $x_{1i}$ with the 15 dimensions unique to the second experiment from that cell after mapping $G_{12}(x_{1i})$.

For combining the measurements from each experiment to be meaningful, the mapped point $G_{12}(x_{1i})$ must correspond accurately to the true point $x_{2i}$. This notion can be captured by taking the correspondence loss function to be the MSE across the 16 dimensions that are shared between the experiments. In other words, the MAGAN should use the shared measurements to match cells between experiments, and then learn the required mapping for all of the measurements that are not shared. Without incorporating this correspondence measure into the model, $x_{1i}$ need not be analogous to $G_{12}(x_{1i})$ in any way, and their information could not be combined.

We evaluate the accuracy of each model's learned correspondence by removing one of the markers measured in both experiments, CD3, from the first experiment. Then, we map points from the first experiment to the second experiment and evaluate how well the discovered CD3 values correspond with the true, held-out CD3 values for each cell from the first experiment.

\begin{figure}
\centering
\includegraphics[width=.5\textwidth]{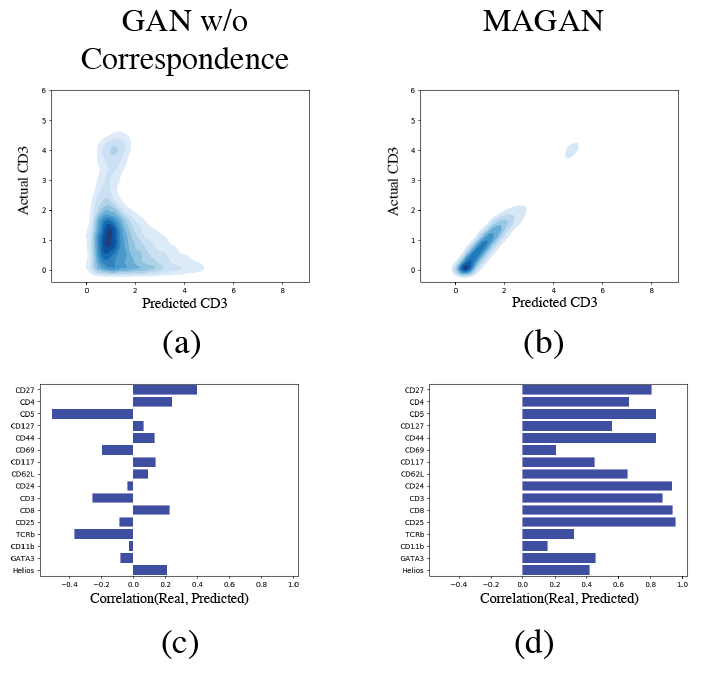}
\caption{Using the MAGAN's correspondence loss, measurements from each experiment can be combined. Their true values are known because they are measured in both experiments. Performing cross-validation by holding each out from the first experiment, we can measure the correlation between the predicted value and the real, correct value.}
\label{fig:paired_cytof}
\end{figure}

Figure \ref{fig:paired_cytof}a shows that the GAN without the correspondence loss finds a manifold superimposition that does not preserve the values of CD3 for each cell accurately. Quantitatively, we can evaluate this with the correlation coefficient between the real, held-out CD3 values and the CD3 values predicted after mapping each point to the other domain. For the GAN without the correspondence loss (Figure \ref{fig:paired_cytof}a), the correlation is -.275, while for the MAGAN (Figure \ref{fig:paired_cytof}b) it is .801. The negative correlation means that without the correspondence loss, the GAN will systematically map cells in one panel to different cells in the other panel.

We perform cross-validation by repeating this test with each of the 16 shared markers in turn for the GAN without correspondence loss (Figure \ref{fig:paired_cytof}c) and the MAGAN (Figure \ref{fig:paired_cytof}d). While some of the markers have more shared information than others and are recovered more accurately, in all cases the correlation is better with the MAGAN.

If we had not measured one of these in the first experiment, we would have been able to use the learned value from the mapping in its place with remarkable accuracy. The MAGAN can powerfully increase the impact of CyTOF experiments by expanding their limited capacity of markers that can be measured at any one time.

\subsection{Correspondence Between CyTOF and scRNA-seq}
To demonstrate the MAGAN aligning manifolds of domains with radically different dimensionality and underlying structure, we use it to find correspondences between CyTOF and scRNA-seq measurements made on the same set of cells. These two types of measurements have advantages and disadvantages, including the throughput, quality, and amount of information acquired from each. Being able to combine their information offers the possibility of getting the best from each and finding insights that might not otherwise be obtainable. In order to do this, though, it is crucial for pointwise correspondences to be accurate, or else features of a data point in the scRNA-seq domain will be ascribed to the incorrect point in the CyTOF domain and the relationships will be meaningless.

To test the MAGAN in this setting, we use a dataset consisting of 2830 measurements, where the dimensionality of each domain is 12 and 12496 for CyTOF and scRNA-seq, respectively \cite{velten2017human}. The scRNA-seq data was normalized with the inverse hyperbolic sine transform and preprocessed with MAGIC \cite{van2017magic}. Here we know the true correspondences of which points in the two domains are the same cell. In this setting we use the semi-supervised correspondence loss and show the impact of providing the pairing of just 10 cells, which can easily be acquired with a few minutes of inspection. This effort is dwarfed by the time and expense involved in acquiring the data from the actual biological experiments themselves, and now substantially improves what we can do with that data.

We evaluate the quality of the correspondences learned with two metrics. First, we calculate the \textit{correspondence error}, or MSE between the true known value $x_{1i} \in X_1$ and the predicted correspondence $G_{21}(x_{2i})$. Table \ref{tab:correspondence_error} shows the correspondence error for the correspondences mapping to and from each domain. With the correspondence loss, the MAGAN cuts the MSE dramatically. 

\begin{table}
\caption{With the MAGAN's correspondence loss, the accuracy of the learned mapping is dramatically improved, as measured by the MSE between the known real point $x$ and the predicted point $G(x)$ after mapping.}
\begin{center}
\begin{tabular}{| c | c | c |}
\hline
\thead{Paired CyTOF \\ \& scRNA-seq}  & \thead{Without \\Correspondence\\Loss} & \thead{With \\Correspondence\\Loss}\\ 
\hline
\thead{MSE($x_1$, $G_{21}(x_2))$} & 99.3 & 22.0\\
\hline
\thead{MSE($x_2$, $G_{12}(x_1))$} & 33.7 & 7.1\\
\hline
\end{tabular}
\end{center}
\label{tab:correspondence_error}
\end{table}

\section{Discussion}
A considerable amount of work has been devoted to GAN architectures in recent years. After the original paper introduced the GAN \cite{goodfellow2014generative}, the difficulty in training them prompted the need for improved training techniques \cite{salimans2016improved}.

Beyond the unsupervised models discussed earlier, other models have tried improving the accuracy of found correspondences with supervision or semi-supervision. These models have included forcing the networks to model specific additional codes or representations and conditioning on external variables such as text \cite{tran2017disentangled, perarnau2016invertible, reed2016generative}.

Other approaches decompose the primary task of the GAN into separate, domain-specific tasks performed sequentially \cite{liu2016coupled, zhang2017stackgan, wang2016generative}. All of these have focused on image domains.

Compared to image domains, relatively little has been done with GANs in biological domains. \cite{shaham2017removal} does not use a GAN, but performs batch correction with neural networks.

We show here how GANs can be used for tasks where the generation of new samples from a given distribution is not the primary goal. In these cases, other terms in the objective function can be used to better match the loss landscape with the task at hand. The MAGAN illustrates one such re-purposing of the GAN architecture that outperforms the existing architectures at finding point-wise correspondences between domains.

\section{Conclusion}
The MAGAN discovers relationships between domains by aligning their manifolds rather than just superimposing them. Crucially, this can be used when one system is measured in two different ways and thus forms two different manifolds. In this case, the point in each manifold for one object in the underlying system are linked. This preserves information at a pointwise (rather than just population aggregate) level.

The MAGAN facilitates integration of datasets from multiple biological modalities. As each type of experiment captures different information with different strengths and weaknesses, combining them makes possible discoveries that could not be found otherwise.

\bibliography{bibliography}

\begin{thebibliography}{24}
\providecommand{\natexlab}[1]{#1}
\providecommand{\url}[1]{#1}
\csname url@samestyle\endcsname
\providecommand{\newblock}{\relax}
\providecommand{\bibinfo}[2]{#2}
\providecommand{\BIBentrySTDinterwordspacing}{\spaceskip=0pt\relax}
\providecommand{\BIBentryALTinterwordstretchfactor}{4}
\providecommand{\BIBentryALTinterwordspacing}{\spaceskip=\fontdimen2\font plus
\BIBentryALTinterwordstretchfactor\fontdimen3\font minus
  \fontdimen4\font\relax}
\providecommand{\BIBforeignlanguage}[2]{{%
\expandafter\ifx\csname l@#1\endcsname\relax
\typeout{** WARNING: IEEEtranN.bst: No hyphenation pattern has been}%
\typeout{** loaded for the language `#1'. Using the pattern for}%
\typeout{** the default language instead.}%
\else
\language=\csname l@#1\endcsname
\fi
#2}}
\providecommand{\BIBdecl}{\relax}
\BIBdecl

\bibitem[Isola et~al.(2016)Isola, Zhu, Zhou, and Efros]{isola2016image}
P.~Isola, J.-Y. Zhu, T.~Zhou, and A.~A. Efros, ``Image-to-image translation
  with conditional adversarial networks,'' \emph{arXiv preprint
  arXiv:1611.07004}, 2016.

\bibitem[Zhu et~al.(2017)Zhu, Park, Isola, and Efros]{zhu2017unpaired}
J.-Y. Zhu, T.~Park, P.~Isola, and A.~A. Efros, ``Unpaired image-to-image
  translation using cycle-consistent adversarial networks,'' \emph{arXiv
  preprint arXiv:1703.10593}, 2017.

\bibitem[Yi et~al.(2017)Yi, Zhang, Tan, and Gong]{yi2017dualgan}
Z.~Yi, H.~Zhang, P.~Tan, and M.~Gong, ``Dualgan: Unsupervised dual learning for
  image-to-image translation,'' \emph{arXiv preprint}, 2017.

\bibitem[Kim et~al.(2017)Kim, Cha, Kim, Lee, and Kim]{kim2017learning}
T.~Kim, M.~Cha, H.~Kim, J.~Lee, and J.~Kim, ``Learning to discover cross-domain
  relations with generative adversarial networks,'' \emph{arXiv preprint
  arXiv:1703.05192}, 2017.

\bibitem[Domingos(2012)]{domingos2012few}
P.~Domingos, ``A few useful things to know about machine learning,''
  \emph{Communications of the ACM}, vol.~55, no.~10, pp. 78--87, 2012.

\bibitem[Park et~al.(2017)Park, Anand, Moniz, Lee, Chakraborty, Choo, Park, and
  Kim]{park2017mmgan}
N.~Park, A.~Anand, J.~R.~A. Moniz, K.~Lee, T.~Chakraborty, J.~Choo, H.~Park,
  and Y.~Kim, ``Mmgan: Manifold matching generative adversarial network for
  generating images,'' \emph{arXiv preprint arXiv:1707.08273}, 2017.

\bibitem[Zhu et~al.(2016)Zhu, Kr{\"a}henb{\"u}hl, Shechtman, and
  Efros]{zhu2016generative}
J.-Y. Zhu, P.~Kr{\"a}henb{\"u}hl, E.~Shechtman, and A.~A. Efros, ``Generative
  visual manipulation on the natural image manifold,'' in \emph{European
  Conference on Computer Vision}.\hskip 1em plus 0.5em minus 0.4em\relax
  Springer, 2016, pp. 597--613.

\bibitem[Bendall et~al.(2012)Bendall, Nolan, Roederer, and
  Chattopadhyay]{bendall2012deep}
S.~C. Bendall, G.~P. Nolan, M.~Roederer, and P.~K. Chattopadhyay, ``A deep
  profiler's guide to cytometry,'' \emph{Trends in immunology}, vol.~33, no.~7,
  pp. 323--332, 2012.

\bibitem[Klein et~al.(2015)Klein, Mazutis, Akartuna, Tallapragada, Veres, Li,
  Peshkin, Weitz, and Kirschner]{klein2015droplet}
A.~M. Klein, L.~Mazutis, I.~Akartuna, N.~Tallapragada, A.~Veres, V.~Li,
  L.~Peshkin, D.~A. Weitz, and M.~W. Kirschner, ``Droplet barcoding for
  single-cell transcriptomics applied to embryonic stem cells,'' \emph{Cell},
  vol. 161, no.~5, pp. 1187--1201, 2015.

\bibitem[Hinton et~al.(1997)Hinton, Dayan, and Revow]{hinton1997modeling}
G.~E. Hinton, P.~Dayan, and M.~Revow, ``Modeling the manifolds of images of
  handwritten digits,'' \emph{IEEE transactions on Neural Networks}, vol.~8,
  no.~1, pp. 65--74, 1997.

\bibitem[Vincent et~al.(2008)Vincent, Larochelle, Bengio, and
  Manzagol]{vincent2008extracting}
P.~Vincent, H.~Larochelle, Y.~Bengio, and P.-A. Manzagol, ``Extracting and
  composing robust features with denoising autoencoders,'' in \emph{Proceedings
  of the 25th international conference on Machine learning}.\hskip 1em plus
  0.5em minus 0.4em\relax ACM, 2008, pp. 1096--1103.

\bibitem[Salimans et~al.(2016)Salimans, Goodfellow, Zaremba, Cheung, Radford,
  and Chen]{salimans2016improved}
T.~Salimans, I.~Goodfellow, W.~Zaremba, V.~Cheung, A.~Radford, and X.~Chen,
  ``Improved techniques for training gans,'' in \emph{Advances in Neural
  Information Processing Systems}, 2016, pp. 2234--2242.

\bibitem[Capra et~al.(1999)Capra, Janeway, Travers, and
  Walport]{capra1999inmunobiology}
J.~D. Capra, C.~A. Janeway, P.~Travers, and M.~Walport, \emph{Inmunobiology:
  the inmune system in health and disease}.\hskip 1em plus 0.5em minus
  0.4em\relax Garland Publishing,, 1999.

\bibitem[Setty et~al.(2016)Setty, Tadmor, Reich-Zeliger, Angel, Salame,
  Kathail, Choi, Bendall, Friedman, and Pe'er]{setty2016wishbone}
M.~Setty, M.~D. Tadmor, S.~Reich-Zeliger, O.~Angel, T.~M. Salame, P.~Kathail,
  K.~Choi, S.~Bendall, N.~Friedman, and D.~Pe'er, ``Wishbone identifies
  bifurcating developmental trajectories from single-cell data,'' \emph{Nature
  biotechnology}, vol.~34, no.~6, p. 637, 2016.

\bibitem[Velten et~al.(2017)Velten, Haas, Raffel, Blaszkiewicz, Islam, Hennig,
  Hirche, Lutz, Buss, Nowak, et~al.]{velten2017human}
L.~Velten, S.~F. Haas, S.~Raffel, S.~Blaszkiewicz, S.~Islam, B.~P. Hennig,
  C.~Hirche, C.~Lutz, E.~C. Buss, D.~Nowak \emph{et~al.}, ``Human
  haematopoietic stem cell lineage commitment is a continuous process,''
  \emph{Nature cell biology}, vol.~19, no.~4, p. 271, 2017.

\bibitem[van Dijk et~al.(2017)van Dijk, Nainys, Sharma, Kathail, Carr, Moon,
  Mazutis, Wolf, Krishnaswamy, and Pe'er]{van2017magic}
D.~van Dijk, J.~Nainys, R.~Sharma, P.~Kathail, A.~J. Carr, K.~R. Moon,
  L.~Mazutis, G.~Wolf, S.~Krishnaswamy, and D.~Pe'er, ``Magic: A
  diffusion-based imputation method reveals gene-gene interactions in
  single-cell rna-sequencing data,'' \emph{BioRxiv}, p. 111591, 2017.

\bibitem[Goodfellow et~al.(2014)Goodfellow, Pouget-Abadie, Mirza, Xu,
  Warde-Farley, Ozair, Courville, and Bengio]{goodfellow2014generative}
I.~Goodfellow, J.~Pouget-Abadie, M.~Mirza, B.~Xu, D.~Warde-Farley, S.~Ozair,
  A.~Courville, and Y.~Bengio, ``Generative adversarial nets,'' in
  \emph{Advances in neural information processing systems}, 2014, pp.
  2672--2680.

\bibitem[Tran et~al.(2017)Tran, Yin, and Liu]{tran2017disentangled}
L.~Tran, X.~Yin, and X.~Liu, ``Disentangled representation learning gan for
  pose-invariant face recognition,'' in \emph{CVPR}, vol.~3, 2017, p.~7.

\bibitem[Perarnau et~al.(2016)Perarnau, van~de Weijer, Raducanu, and
  {\'A}lvarez]{perarnau2016invertible}
G.~Perarnau, J.~van~de Weijer, B.~Raducanu, and J.~M. {\'A}lvarez, ``Invertible
  conditional gans for image editing,'' \emph{arXiv preprint arXiv:1611.06355},
  2016.

\bibitem[Reed et~al.(2016)Reed, Akata, Yan, Logeswaran, Schiele, and
  Lee]{reed2016generative}
S.~Reed, Z.~Akata, X.~Yan, L.~Logeswaran, B.~Schiele, and H.~Lee, ``Generative
  adversarial text to image synthesis,'' \emph{arXiv preprint
  arXiv:1605.05396}, 2016.

\bibitem[Liu and Tuzel(2016)]{liu2016coupled}
M.-Y. Liu and O.~Tuzel, ``Coupled generative adversarial networks,'' in
  \emph{Advances in neural information processing systems}, 2016, pp. 469--477.

\bibitem[Zhang et~al.(2017)Zhang, Xu, Li, Zhang, Huang, Wang, and
  Metaxas]{zhang2017stackgan}
H.~Zhang, T.~Xu, H.~Li, S.~Zhang, X.~Huang, X.~Wang, and D.~Metaxas,
  ``Stackgan: Text to photo-realistic image synthesis with stacked generative
  adversarial networks,'' in \emph{IEEE Int. Conf. Comput. Vision (ICCV)},
  2017, pp. 5907--5915.

\bibitem[Wang and Gupta(2016)]{wang2016generative}
X.~Wang and A.~Gupta, ``Generative image modeling using style and structure
  adversarial networks,'' in \emph{European Conference on Computer
  Vision}.\hskip 1em plus 0.5em minus 0.4em\relax Springer, 2016, pp. 318--335.

\bibitem[Shaham et~al.(2017)Shaham, Stanton, Zhao, Li, Raddassi, Montgomery,
  and Kluger]{shaham2017removal}
U.~Shaham, K.~P. Stanton, J.~Zhao, H.~Li, K.~Raddassi, R.~Montgomery, and
  Y.~Kluger, ``Removal of batch effects using distribution-matching residual
  networks,'' \emph{Bioinformatics}, p. btx196, 2017.

\end{thebibliography}
\bibliographystyle{IEEEtranN}

\end{document}